\documentclass[
]{ceurart}

\usepackage{caption}
\usepackage{subcaption}

\usepackage{adjustbox}
\usepackage{listings}
\begin{document}

\copyrightyear{2023}
\copyrightclause{Copyright for this paper by its authors.
  Use permitted under Creative Commons License Attribution 4.0
  International (CC BY 4.0).}

\conference{IberLEF 2023, September 2023, Jaén, Spain}
\title{Early Detection of Depression and Eating Disorders in Spanish: UNSL at MentalRiskES 2023}


\author[1,2]{Horacio Thompson}[
email=hjthompson@unsl.edu.ar,
]

\author[1]{Marcelo Errecalde}[%
email=merreca@unsl.edu.ar,
]

\address[1]{Universidad Nacional de San Luis (UNSL), Ejército de Los Andes 950, San Luis, C.P. 5700, Argentina}
\address[2]{Consejo Nacional de Investigaciones Científicas y Técnicas (CONICET), San Luis, Argentina}


\begin{abstract}
MentalRiskES is a novel challenge that proposes to solve problems related to early risk detection for the Spanish language. The objective is to detect, as soon as possible, Telegram users who show signs of mental disorders considering different tasks. Task 1 involved the users' detection of eating disorders, Task 2 focused on depression detection, and Task 3 aimed at detecting an unknown disorder. These tasks were divided into subtasks, each one defining a resolution approach. \\
Our research group participated in subtask A for Tasks 1 and 2: a binary classification problem that evaluated whether the users were positive or negative. To solve these tasks, we proposed models based on Transformers followed by a decision policy according to criteria defined by an early detection framework. One of the models presented an extended vocabulary with important words for each task to be solved. In addition, we applied a decision policy based on the history of predictions that the model performs during user evaluation. \\
For Tasks 1 and 2, we obtained the second-best performance according to rankings based on classification and latency, demonstrating the effectiveness and consistency of our approaches for solving early detection problems in the Spanish language.
\end{abstract}


\maketitle

\section{Introduction}
According to the World Health Organization, one in every eight people worldwide suffers a mental disorder. Anxiety, depression, bipolar disorder, and eating behavior disorders are the most frequent \cite{charlson2019new}. Different social networks have become mass media chosen by people to share information and express their emotions. Several studies show the relationship between the use of social networks and mental disorders \cite{aliverdi2022social, maulik2010effect, martinez2022redes}. Therefore, there is a growing interest in the early identification of users suffering from these disorders to provide them with appropriate help. Evaluation conferences such as CLEF eRisk have promoted research groups to solve early detection challenges considering different domains \cite{losada2016erisk, losada2017erisk, losada2018overview, losada2019overview, losada2020erisk, parapar2021overview, parapar2022overview}. However, there is currently no challenge of these characteristics for Spanish, highlighting the urgent need to promote initiatives in this language.

MentalRiskES is a novel challenge that proposes to solve problems of early risk detection of mental disorders in Spanish \cite{MentalRiskES2023}. In this first edition, three tasks were defined with the same objective: to detect Telegram users who show signs of mental disorders as early as possible. Task 1 consisted of the detection of users with eating disorders, Task 2 was related to depression detection, and Task 3 to the detection of an unknown disorder. Each task was divided into subtasks for solving the problem considering different approaches: 

\vspace{0.1cm} \noindent
\textbf{Binary classification (subtask A):} To decide whether or not a user suffers from a mental disorder by considering the positive and negative classes.\\
\textbf{Simple regression (subtask B):} To provide an affectation probability on positive and negative classes.\\
\textbf{Multi-class classification (subtask C):} To decide whether a user suffers from a mental disorder and evaluate their attitude towards it by considering additional classes.\\
\textbf{Multi-output regression (subtask D):} To provide a confidence probability for the additional classes. 

\vspace{0.1cm} \noindent
Subtasks A and B were defined for Tasks 1, 2, and 3, while subtasks C and D were also included for Task 2.

Early risk detection can be analyzed as a multi-objective problem, where the challenge is to find an adequate balance between the precision in identifying risky users and the minimum time required for that decision to be reliable. 
Our research group achieved notable results in the 2021 \cite{loyola2021unsl}, 2022 \cite{loyola2022unsl}, and 2023 (article currently under review) editions of the CLEF eRisk. In these last two editions, we used an early detection framework \cite{loyola2018learning} which defines that it is necessary to consider two components: one dedicated to solving a user classification problem (classification with partial information or CPI), and the other involves a decision policy to decide when to stop evaluating a user (deciding the moment of classification or DMC). In particular, we applied the framework by using a BERT model \cite{devlin2018bert} with an extended vocabulary (CPI component) and a decision policy based on a historic rule (DMC component). In this first edition of MentalRiskES, we participated in Tasks 1 and 2 according to subtask A. Following our Transformers-based approaches, we used the BETO model \cite{CaneteCFP2020}, a variant of BERT that was trained on large Spanish corpora, and we adjusted the historic rule according to the tasks to be solved.

The present work describes the approaches used by our research group to solve Tasks 1 and 2. Section \ref{sec:resolution} details the datasets, classification models, and decision policies applied. Section \ref{sec:results} shows the results obtained in both tasks and Section \ref{sec:conclusion} presents the conclusions and future works.

\section{Resolution Method}
\label{sec:resolution}

The challenge was divided into two stages: a \emph{training stage}, where the participants experimented with data provided by the Organizers, and a \emph{test stage}, where a client application interacted with a server, defining an early environment. This last process was carried out in rounds, during which the client requested the next post of users and, according to the number of predictive models, evaluated them and returned a response to the server.

\subsection{Datasets}
Three corpora were available to solve Tasks 1 and 2, as shown in Table \ref{tab:datasets}. The \emph{Train} and \emph{Trial} corpora were available for the participants to implement their proposals. The \emph{Trial} corpora were proposed to test the connection between the client application and the server. The \emph{Test} corpora were used for the Organizers to evaluate the participating models. 
It should be noted that, for both tasks, the number of corpora samples is limited. In the training stage, 185 samples between \emph{Train} and \emph{Trial} were available, while in the test stage, 150 users were evaluated. Furthermore, in contrast to what typically occurs in these classification problems, the classes exhibit a relatively balanced distribution, as evidenced by the number of positive and negative users. On the other hand, for Tasks 1 and 2, the median number of posts per user is approximately 21 and 31, respectively. This fact is relevant because a model with an acceptable performance should finish the evaluation of the users in a smaller number of posts. The maximum number of posts per user in the \emph{Test} corpus indicates the number of total rounds that the test stage had: 50 for Task 1 and 100 for Task 2. Finally, it is observed that the posts were relatively short (between 8 and 9 words per post for each task).

\begin{table}[h]
\caption{Details of the corpora of Tasks 1 and 2. The number of users (total, positives, and negatives) and the number of posts in each corpus are reported. The median, minimum, and maximum number of posts per user and words per post in each corpus are detailed.}
\label{tab:datasets}
\begin{tabular}{|c|c|ccc|c|ccc|ccc|}
\hline
\multicolumn{1}{|l|}{\multirow{2}{*}{\textbf{}}} & \multirow{2}{*}{\textbf{Corpus}} & \multicolumn{3}{c|}{\textbf{\#Users}}                                                  & \multirow{2}{*}{\textbf{\#Posts}} & \multicolumn{3}{c|}{\textbf{\#Posts per user}}                                       & \multicolumn{3}{c|}{\textbf{\#Words per post}}                                       \\ \cline{3-5} \cline{7-12} 
\multicolumn{1}{|l|}{}                           &                                  & \multicolumn{1}{c|}{\textbf{Total}} & \multicolumn{1}{c|}{\textbf{Pos}} & \textbf{Neg} &                                   & \multicolumn{1}{c|}{\textbf{Med}} & \multicolumn{1}{c|}{\textbf{Min}} & \textbf{Max} & \multicolumn{1}{c|}{\textbf{Med}} & \multicolumn{1}{c|}{\textbf{Min}} & \textbf{Max} \\ \hline
\multirow{3}{*}{\textbf{Task 1}}                 & Train                            & \multicolumn{1}{c|}{175}            & \multicolumn{1}{c|}{74}           & 101          & 5931                              & \multicolumn{1}{c|}{35.0}         & \multicolumn{1}{c|}{11}           & 50           & \multicolumn{1}{c|}{9.0}          & \multicolumn{1}{c|}{2}            & 899          \\
                                                 & Trial                            & \multicolumn{1}{c|}{10}             & \multicolumn{1}{c|}{5}            & 5            & 389                               & \multicolumn{1}{c|}{48.5}         & \multicolumn{1}{c|}{18}           & 50           & \multicolumn{1}{c|}{9.0}          & \multicolumn{1}{c|}{3}            & 753          \\
                                                 & Test                             & \multicolumn{1}{c|}{150}            & \multicolumn{1}{c|}{64}           & 86           & 4179                              & \multicolumn{1}{c|}{21.5}         & \multicolumn{1}{c|}{11}           & 50           & \multicolumn{1}{c|}{9.0}          & \multicolumn{1}{c|}{2}            & 894          \\ \hline
\multirow{3}{*}{\textbf{Task 2}}                 & Train                            & \multicolumn{1}{c|}{175}            & \multicolumn{1}{c|}{94}           & 81           & 6248                              & \multicolumn{1}{c|}{26.0}         & \multicolumn{1}{c|}{11}           & 100          & \multicolumn{1}{c|}{9.0}          & \multicolumn{1}{c|}{1}            & 783          \\
                                                 & Trial                            & \multicolumn{1}{c|}{10}             & \multicolumn{1}{c|}{6}            & 4            & 624                               & \multicolumn{1}{c|}{68.0}         & \multicolumn{1}{c|}{11}           & 100          & \multicolumn{1}{c|}{9.0}          & \multicolumn{1}{c|}{3}            & 201          \\
                                                 & Test                             & \multicolumn{1}{c|}{149}            & \multicolumn{1}{c|}{68}           & 81           & 5164                              & \multicolumn{1}{c|}{31.0}         & \multicolumn{1}{c|}{11}           & 100          & \multicolumn{1}{c|}{8.0}          & \multicolumn{1}{c|}{1}            & 368          \\ \hline
\end{tabular}
\end{table}

\subsection{CPI components: Models}
\vspace{0.1cm} \noindent
\textbf{Training set.} Due to the limited data available, we augmented the number of samples. For each user, we divided the list of posts into three equal parts according to the list length. Each portion was labeled using the user's label and added to the training set.
In this way, we obtained approximately 500 new samples to train the models. Besides, this allowed the models to be trained considering different contexts of the users' history and trying to overcome the limitation of BERT architectures that only admit 512 input tokens. Then, each model was trained and validated using an 85/15 split of the \emph{Train} corpus with added samples.

\vspace{0.1cm} \noindent
\textbf{Preprocessing.} Some preprocessing actions were performed before the \emph{fine-tuning} process. Characters were converted to lowercase, while Unicode and HTML codes were transformed into their corresponding symbols. Web pages and numbers were replaced by the \emph{weblink} and \emph{number} tokens, respectively. Repeated words and spaces were also removed. 
\newpage
\vspace{0.1cm} \noindent
\textbf{Classifier type.} We used the BETO model (version: \emph{dccuchile/bert-base-spanish-wwm-uncased}), applying the \emph{fine-tuning} process to adjust it to each task. Different hyperparameters were considered, and a scheduler was used to automatically adjust the learning rate during \emph{fine-tuning}, improving the model convergence. For Tasks 1 and 2, we presented two proposals:
\begin{itemize}
    \item 
    \textbf{Classic BETO model}. We imported the pre-trained model and applied the \emph{fine-tuning} process. It was a baseline model.
    \item 
    \textbf{BETO model with an extended vocabulary}. 
    Important words were added according to the task to be solved. They were extracted from an external model known as SS3 \cite{burdisso2019text}. We trained SS3 to classify users on the available corpora, and we selected the best words according to the confidence values on the positive class. For Task 1, \emph{ayuno} (fasting), \emph{cals} (calories acronym), \emph{atracones} (binge eating), and for Task 2, \emph{decepcionada} (disappointed), \emph{suicidarme} (to commit suicide), and \emph{daño} (damage) are some examples of important words. We evaluated the number of words added to each model in a range of 5 to 50 by considering the validation performances.
\end{itemize} 
Finally, the best CPI model for each proposal was chosen according to the F1 metric over the positive class (F1+). Table \ref{tab:hyperparams} shows a summary of the hyperparameters selected for each task.

\begin{table}[h]
\caption{Hyperparameters of the models presented for Tasks 1 and 2.}
\label{tab:hyperparams}
\begin{adjustbox}{width=\textwidth}
\begin{tabular}{|c|c|c|c|c|c|c|c|c|}
\hline
\multicolumn{1}{|l|}{\textbf{}}  & \textbf{Team\#Run} & \textbf{\begin{tabular}[c]{@{}c@{}}Model\\ type\end{tabular}}       & \textbf{\begin{tabular}[c]{@{}c@{}}\#Batch\\ size\end{tabular}} & \textbf{\begin{tabular}[c]{@{}c@{}}Learning\\ rate\end{tabular}} & \textbf{\#Epochs} & \textbf{\begin{tabular}[c]{@{}c@{}}\#Added\\ words\end{tabular}} & \textbf{Optimizer}      & \multicolumn{1}{l|}{\textbf{Scheduler}}                                              \\ \hline
\multirow{2}{*}{\textbf{Task 1}} & \textbf{UNSL\#0}   & Classic BETO                                                        & 8                                                               & 3E-5                                                             & 5                 & -                                                                & \multirow{4}{*}{AdamW} & \multirow{4}{*}{\begin{tabular}[c]{@{}c@{}}Linear\\ Scheduler\\ Warmup\end{tabular}} \\ \cline{2-7}
                                 & \textbf{UNSL\#1}   & \begin{tabular}[c]{@{}c@{}}BETO with\\ extended vocabulary\end{tabular} & 8                                                               & 2E-5                                                             & 3                 & 25                                                               &                        &                                                                                      \\ \cline{1-7}
\multirow{2}{*}{\textbf{Task 2}} & \textbf{UNSL\#0}   & Classic BETO                                                        & 8                                                               & 2E-5                                                             & 5                 & -                                                                &                        &                                                                                      \\ \cline{2-7}
                                 & \textbf{UNSL\#1}   & \begin{tabular}[c]{@{}c@{}}BETO with\\ extended vocabulary\end{tabular} & 8                                                               & 2E-5                                                             & 5                 & 25                                                               &                        &                                                                                      \\ \hline
\end{tabular}
\end{adjustbox}
\end{table}


\subsection{DMC component: Decision Policy}
The next step was to find the best decision policy for each task using a mock server (available in: \url{https://github.com/jmloyola/erisk\_mock\_server}). This tool simulates the eRisk challenge through the rounds of posts and answers submissions, and it allows the calculation of the final results according to metrics based on decision and ranking. 
It was helpful since the performance of CPI models can drastically change when evaluated in an early environment. A client application was defined to manage the interaction with the server. When it receives a round of posts, the system preprocesses the writings, invokes the predictive models (CPI), and applies a decision policy (DMC). To take advantage of the 512 input tokens that the BETO architecture admits, the application uses the last \emph{N} = 10 posts (posts window), linking the current with previous posts. With the mock server, the client application, and the predictive models, different decision policies were evaluated using the F1+, ERDE-5, ERDE-50, and latency-weighted F1 metrics. It should be noted that the client application was also used in the test stage of MentalRiskES.
\newpage
\vspace{0.1cm} \noindent
\textbf{Decision policy: Historic rule} 

\vspace{0.1cm} \noindent
``\emph{If the current prediction and last M predictions exceed T times a Threshold, the client application must issue a risky user alarm; otherwise, it is necessary to continue the user evaluation}''. \\
The parameter \emph{M} is the number of past predictions that the rule considers, \emph{T} is the tolerance, i.e., how many predictions can exceed the \emph{Threshold} before issuing an alarm, and \emph{Threshold} is the limit probability to predict a user as positive. In addition, the rule has the \emph{min\_delay} parameter, which defines the moment when it will start to apply. Table \ref{tab:hr-params} shows the best parameters for each task, which were found by evaluating the models with the mock server on the \emph{Trial} corpus. 
\begin{table}[h]
\caption{Best parameters to define the decision policy based on the historic rule. The rule was applied for the UNSL\#0 and UNSL\#1 models in both tasks. \emph{M}: number of past predictions; \emph{T}: tolerance; \emph{Threshold}: the limit probability to predict a positive user; \emph{min\_delay}: the moment when the rule will apply.}
\label{tab:hr-params}
\begin{tabular}{|c|c|c|c|c|}
\hline
                & \textbf{\begin{tabular}[c]{@{}c@{}}M\end{tabular}} & \textbf{\begin{tabular}[c]{@{}c@{}}T \end{tabular}} & \textbf{\begin{tabular}[c]{@{}c@{}}Threshold\end{tabular}} & \textbf{\begin{tabular}[c]{@{}c@{}}min\_delay\end{tabular}} \\ \hline
\textbf{HistoricRule\_T1} & all predictions                                                       & \begin{tabular}[c]{@{}c@{}}5 \\ predictions\end{tabular}         & 0.7                                                                                 & \begin{tabular}[c]{@{}c@{}}after 5\\ predictions\end{tabular}             \\ \hline
\textbf{HistoricRule\_T2} & all predictions                                                       & \begin{tabular}[c]{@{}c@{}}10\\ predictions\end{tabular}         & 0.7                                                                            & \begin{tabular}[c]{@{}c@{}}after 10\\ predictions\end{tabular}             \\ \hline
\end{tabular}   
\end{table}

\vspace{0.1cm} \noindent
In summary, the final models to solve Tasks 1 and 2 were: 
\begin{table}[h]
\centering
\begin{tabular}{| l |}
\hline
\textbf{Task 1: Eating disorders}\\
\hspace{0.5cm} \textbf{UNSL\#0:} Classic BETO (CPI) + HistoricRule\_T1 (DMC) \\
\hspace{0.5cm} \textbf{UNSL\#1:} BETO with an extended vocabulary (CPI) + HistoricRule\_T1 (DMC) \\
\textbf{Task 2: Depression}\\
\hspace{0.5cm} \textbf{UNSL\#0:} Classic BETO (CPI) + HistoricRule\_T2 (DMC) \\
\hspace{0.5cm} \textbf{UNSL\#1:} BETO with an extended vocabulary (CPI) + HistoricRule\_T2 (DMC) \\
\hline
\end{tabular}
\end{table}

\section{Results}
\label{sec:results}
The Organizers evaluated the teams considering metrics based on classification and latency for subtask A of Tasks 1 and 2. The first metrics evaluate the models according to classification performance, while the second ones penalize performance considering the number of posts required to detect positive users. The Organizers published a results report with team rankings ordered according to the Macro-F1 (classification-based evaluation) and ERDE-30 (latency-based evaluation) metrics.

\subsection{Task 1 - Subtask A}

Table \ref{tab:classification-t1} shows the results obtained considering the classification metrics. The models with the best Macro-F1 were CIMAT-NLP-GTO\#0 with 0.966, followed by UMUTeam\#0 (0.918) and UNSL\#1 (0.913). Considering the mean and median values among all the proposals (in total, 25), the three models showed excellent classification performance. For its part, UNSL\#0 obtained 0.751, a similar performance to the teams' average. 
According to the latency metrics (Table \ref{tab:latency-t1}), the best ERDE-30 was obtained by CIMAT-NLP-GTO\#0 with 0.018, followed by UNSL\#1 (0.045) and CIMAT-NLP-GTO\#1 (0.065). The best latency-weighted F1 results were achieved by CIMAT-NLP-GTO\#0 (0.863), BaseLine-RobertaLarge\#1 (0.792), and UNSL\#1 (0.776), while the best ERDE-5 was obtained by BaseLine-RobertaLarge\#1 with 0.163. The UNSL\#0 model achieved a better ERDE-30 than the mean and median of the proposals.
In summary, the most outstanding models for Task 1 were CIMAT-NLP-GTO\#0 and UNSL\#1, achieving notable performance in classification and latency.

\begin{table}[h]
\caption{Classification-based evaluation for Task 1 (subtask A). The best team according to the Accuracy, Macro-P, Macro-R, and Macro-F1 metrics is shown (values in bold), as well as the \emph{mean} and \emph{median} values of the results report for MentalRiskES 2023. The second and third-best teams are also included.}
\label{tab:classification-t1}
\begin{tabular}{clcccc}
\hline
\textbf{Ranking} & \textbf{Model\#Run} & \textbf{Accuracy} & \textbf{\begin{tabular}[c]{@{}c@{}}Macro\\ P\end{tabular}} & \textbf{\begin{tabular}[c]{@{}c@{}}Macro\\ R\end{tabular}} & \textbf{\begin{tabular}[c]{@{}c@{}}Macro\\ F1\end{tabular}} \\ \hline
1             & CIMAT-NLP-GTO\#0    & \textbf{0.967}    & \textbf{0.964}                                             & \textbf{0.969}                                             & \textbf{0.966}                                              \\
2             & UMUTeam\#0          & 0.920             & 0.922                                                      & 0.914                                                      & 0.918                                                       \\
3             & UNSL\#1             & 0.913             & 0.912                                                      & 0.920                                                      & 0.913                                                       \\ \hline
15            & UNSL\#0             & 0.753             & 0.817                                                      & 0.785                                                      & 0.751                                                       \\ \hline
\multicolumn{2}{c}{\textit{Mean}}   & 0.765             & 0.820                                                      & 0.786                                                      & 0.750                                                       \\
\multicolumn{2}{c}{\textit{Median}} & 0.810             & 0.830                                                      & 0.829                                                      & 0.810                                                       \\ \hline
\end{tabular}
\end{table}

\begin{table}[h]
\caption{Latency-based evaluation for Task 1 (subtask A). The best teams according to the ERDE-5, ERDE-30, and latency-weighted F1 metrics are shown (values in bold), as well as the \emph{mean} and \emph{median} values of the results report for MentalRiskES 2023. The second and third-best teams are also included.}
\label{tab:latency-t1}
\begin{adjustbox}{width=\textwidth}
\begin{tabular}{clccccc}
\hline
\textbf{Ranking} & \textbf{Model\#Run}       & \textbf{ERDE-5} & \textbf{ERDE-30} & \textbf{latencyTP} & \textbf{speed} & \textbf{latency-weighted F1} \\ \hline
1             & CIMAT-NLP-GTO\#0          & 0.334          & \textbf{0.018}  & 6                  & 0.898          & \textbf{0.863}    \\
2             & UNSL\#1                   & 0.433          & 0.045           & 8                  & 0.857          & 0.776             \\
3             & CIMAT-NLP-GTO\#1          & 0.379          & 0.065           & 6                  & 0.898          & 0.761             \\ \hline
10            & BaseLine-RobertaLarge\#1 & \textbf{0.163} & 0.099           & 2                  & 0.979          & 0.792             \\
11            & UNSL\#0                   & 0.502          & 0.105           & 8                  & 0.867          & 0.673             \\ \hline
\multicolumn{2}{c}{\textit{Mean}}         & 0.322          & 0.122           & 6                  & 0.909          & 0.707             \\
\multicolumn{2}{c}{\textit{Median}}       & 0.306          & 0.112           & 4                  & 0.938          & 0.704             \\ \hline
\end{tabular}
\end{adjustbox}  
\end{table}

\subsection{Task 2 - Subtask A}
Table \ref{tab:classification-t2} shows the classification performances. Considering the Macro-F1 metric, the UMUTeam\#0, UNSL\#1, and UNSL\#0 models obtained the best results, which were very similar to each other. However, UNSL\#1 achieved the best Macro-P and Macro-R. These models notably outperformed the mean and median values among all the proposals (in total, 33). According to the latency metrics (Table \ref{tab:latency-T2}), the SINAI-SELA\#0 model achieved the best ERDE-30 (0.140), followed by UNSL\#1 (0.148) and BaseLine-Deberta\#0 (0.153). Regarding the latency-weighted F1 metric, SINAI-SELA\#0 also obtained the best performance, while the best ERDE-5 was obtained by VICOM-nlp\#2 (0.275). On the other hand, the UNSL\#0 model achieved a higher performance than the mean and median among all the teams for the ERDE-30 metric.
In summary, the best models for this task were SINAI-SELA\#0 and UNSL\#1. Our model achieved the best classification results and remarkable latency performance with the second-best ERDE-30.

\begin{table}[h]
\caption{Classification-based evaluation results for Task 2 (subtask A). The best teams according to the Accuracy, Macro-P, Macro-R, and Macro-F1 metrics are shown (values in bold), as well as the \emph{mean} and \emph{median} values of the results report for MentalRiskES 2023. The SINAI-SELA\#0 model is also included due to its results on latency-based metrics.}
\label{tab:classification-t2}
\begin{tabular}{clcccc}
\hline
\textbf{Ranking} & \textbf{Model\#Run} & \textbf{Accuracy} & \textbf{Macro-P} & \textbf{Macro-R} & \textbf{Macro-F1} \\ \hline
1             & UMUTeam\#0          & \textbf{0.738}    & 0.756            & 0.749            & \textbf{0.737}    \\
2             & UNSL\#1             & \textbf{0.738}    & \textbf{0.791}   & \textbf{0.756}   & 0.733             \\
3             & UNSL\#0             & 0.732             & 0.752            & 0.742            & 0.731             \\ \hline
5             & SINAI-SELA\#0       & 0.725             & 0.775            & 0.742            & 0.720             \\ \hline
\multicolumn{2}{c}{\textit{Mean}}   & 0.617             & 0.710            & 0.637            & 0.579             \\
\multicolumn{2}{c}{\textit{Median}} & 0.631             & 0.731            & 0.658            & 0.616             \\ \hline
\end{tabular}
\end{table}

\begin{table}[h]
\caption{Latency-based evaluation results for Task 2 (subtask A). The best teams according to the ERDE-5, ERDE-30, and latency-weighted F1 are shown (values in bold). The \emph{mean} and \emph{median} values of the results report for MentalRiskES 2023 are shown. The second and third-best teams are also included.}
\label{tab:latency-T2}
\begin{tabular}{clccccc}
\hline
\textbf{Ranking} & \textbf{Model\#Run}   & \textbf{ERDE-5} & \textbf{ERDE-30} & \textbf{latencyTP} & \textbf{speed} & \textbf{\begin{tabular}[c]{@{}c@{}}latency-weighted\\ F1\end{tabular}} \\ \hline
1             & SINAI-SELA\#0         & 0.395           & \textbf{0.140}   & 4                  & 0.951          & \textbf{0.720}                                                         \\
2             & UNSL\#1               & 0.567           & 0.148            & 14                 & 0.791          & 0.609                                                                  \\
3             & BaseLine-Deberta\#0 & 0.303           & 0.153            & 2                  & 0.984          & 0.719                                                                  \\ \hline
8             & VICOM-nlp\#2          & \textbf{0.275}  & 0.173            & 2                  & 0.984          & 0.706                                                                  \\
14            & UNSL\#0               & 0.551           & 0.188            & 14                 & 0.791          & 0.591                                                                  \\ \hline
\multicolumn{2}{c}{\textit{Mean}}     & 0.383           & 0.232            & 8                  & 0.902          & 0.599                                                                  \\
\multicolumn{2}{c}{\textit{Median}}   & 0.362           & 0.205            & 3                  & 0.967          & 0.627                                                                  \\ \hline
\end{tabular}
\end{table}

Finally, the performance of our proposals in terms of efficiency metrics for Tasks 1 and 2 is shown in Table \ref{tab:carbon-emissions}. It is observed that UNSL\#1 and UNSL\#0 outperformed the mean among all the proposals, demonstrating the capability to solve both tasks while minimizing resource requirements and reducing environmental impact.

\begin{table}[h]
\caption{Efficiency metrics for Tasks 1 and 2. The \emph{mean} among all proposals for each metric is shown. }
\label{tab:carbon-emissions}
\resizebox{\textwidth}{!}{%
\begin{tabular}{cccccccccccrr}
\hline
\multicolumn{1}{l}{\textbf{Task}} & \multicolumn{1}{l}{\textbf{Model\#Run}} & \textbf{\begin{tabular}[c]{@{}c@{}}Duration\\ (min)\end{tabular}} & \textbf{Emissions}        & \textbf{\begin{tabular}[c]{@{}c@{}}CPU\\ energy\end{tabular}} & \textbf{\begin{tabular}[c]{@{}c@{}}GPU\\ energy\end{tabular}} & \textbf{\begin{tabular}[c]{@{}c@{}}RAM\\ energy\end{tabular}} & \textbf{\begin{tabular}[c]{@{}c@{}}Consumed\\ energy\end{tabular}} & \textbf{\begin{tabular}[c]{@{}c@{}}CPU\\ count\end{tabular}} & \textbf{\begin{tabular}[c]{@{}c@{}}GPU\\ count\end{tabular}} & \textbf{\begin{tabular}[c]{@{}c@{}}RAM\\ total size\end{tabular}} & \multicolumn{1}{c}{\textbf{\begin{tabular}[c]{@{}c@{}}CPU\\ model\end{tabular}}}                      & \multicolumn{1}{c}{\textbf{\begin{tabular}[c]{@{}c@{}}GPU\\ model\end{tabular}}}       \\ \hline \rule{0pt}{1em}
\multirow{4}{*}{\textbf{1}}       & \textbf{UNSL\#1}                        & 4.644                                                             & 2.78E-05                  & 6.13E-05                                                      & 0                                                             & 1.34E-06                                                      & 6.26E-05                                                           & 16                                                           & 1                                                            & 23.545                                                            & \multirow{3}{*}{\begin{tabular}[c]{@{}r@{}}AMD Ryzen 7 \\ 1700X Eight-Core \\ Processor\end{tabular}} & \multirow{3}{*}{\begin{tabular}[c]{@{}r@{}}1 x \\ GeForce GTX \\ 1080 Ti\end{tabular}} \\
                                  & \multirow{2}{*}{\textbf{UNSL\#0}}       & \multirow{2}{*}{4.631}                                            & \multirow{2}{*}{2.77E-05} & \multirow{2}{*}{6.11E-05}                                     & \multirow{2}{*}{0}                                            & \multirow{2}{*}{1.34E-06}                                     & \multirow{2}{*}{6.24E-05}                                          & \multirow{2}{*}{16}                                          & \multirow{2}{*}{1}                                           & \multirow{2}{*}{23.545}                                           &                                                                                                       &                                                                                        \\
                                  &                                         &                                                                   &                           &                                                               &                                                               &                                                               &                                                                    &                                                              &                                                              &                                                                   &                                                                                                       &                                                                                        \\ \rule{0pt}{1.25em}
                                  & \textit{Mean}                           & 50.840                                                            & 38.45E-05                 & 33.66E-05                                                     & 48.9E-05                                                      & 4.66E-06                                                      & 83.02E-05                                                          & 38                                                           & 4                                                            & 164.198                                                           & \multicolumn{1}{c}{-}                                                                                 & \multicolumn{1}{c}{-}                                                                  \\ \hline \rule{0pt}{1em}
\multirow{4}{*}{\textbf{2}}       & \textbf{UNSL\#1}                        & 3.349                                                             & 2.01E-05                  & 4.42E-05                                                      & 0                                                             & 9.98E-07                                                      & 4.52E-05                                                           & 16                                                           & 1                                                            & 23.545                                                            & \multirow{3}{*}{\begin{tabular}[c]{@{}r@{}}AMD Ryzen 7 \\ 1700X Eight-Core \\ Processor\end{tabular}} & \multirow{3}{*}{\begin{tabular}[c]{@{}r@{}}1 x \\ GeForce GTX \\ 1080 Ti\end{tabular}} \\
                                  & \multirow{2}{*}{\textbf{UNSL\#0}}       & \multirow{2}{*}{3.347}                                            & \multirow{2}{*}{2.01E-05} & \multirow{2}{*}{4.42E-05}                                     & \multirow{2}{*}{0}                                            & \multirow{2}{*}{9.98E-07}                                     & \multirow{2}{*}{4.51E-05}                                          & \multirow{2}{*}{16}                                          & \multirow{2}{*}{1}                                           & \multirow{2}{*}{23.545}                                           &                                                                                                       &                                                                                        \\
                                  &                                         &                                                                   &                           &                                                               &                                                               &                                                               &                                                                    &                                                              &                                                              &                                                                   &                                                                                                       &                                                                                        \\ \rule{0pt}{1.25em}
                                  & \textit{Mean}                           & 34.704                                                            & 50.37E-05                 & 38.78E-05                                                     & 129.97E-05                                                    & 1.17E-05                                                      & 169.92E-05                                                         & 29                                                           & 3                                                            & 123.458                                                           & \multicolumn{1}{c}{-}                                                                                 & \multicolumn{1}{c}{-}                                                                  \\ \hline
\end{tabular}%
}
\end{table}

\subsection{Error analysis}

\begin{figure}[h]
    \centering
    \begin{subfigure}{\textwidth}
        \centering
        \includegraphics[width=0.95\textwidth]{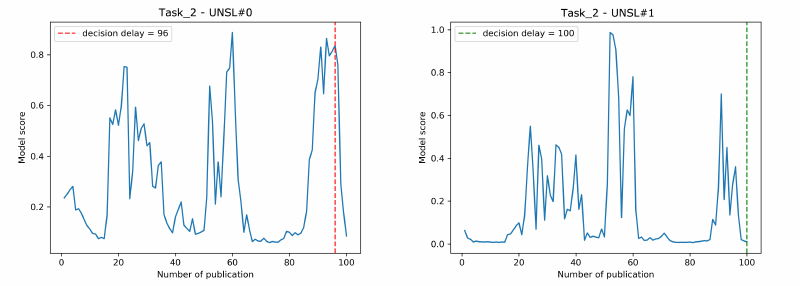}
        \caption{Evaluation of a negative user. UNSL\#0 generated a false positive, while UNSL\#1 classified the user correctly.}
        \label{fig:neg-user}
    \end{subfigure}
    \begin{subfigure}{\textwidth}
        \centering
        \includegraphics[width=0.95\textwidth]{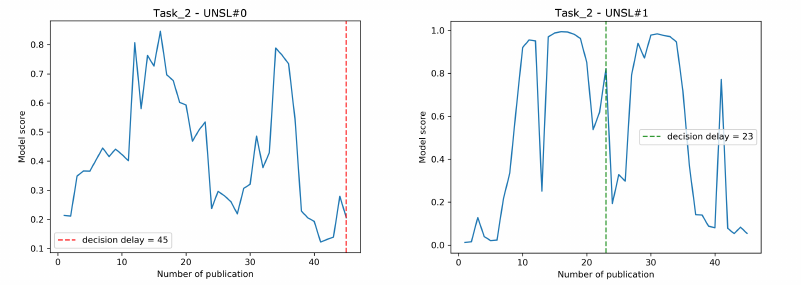}
        \caption{Evaluation of a positive user. UNSL\#0 generated a false negative, while UNSL\#1 classified the user correctly.}
        \label{fig:pos-user}
    \end{subfigure}
    \caption{Comparison between UNSL\#0 and UNSL\#1 to evaluate users of Task 2. The graphics show the evaluation of each model on a negative (a) and positive (b) user. The rounds of posts (\emph{number of publication}) are observed on the $x$-axis, and the model probability at each prediction instance is on the $y$-axis. The dashed lines show the moment of the model's final decision: green (correct prediction) and red (incorrect prediction).}
    \label{fig:eval-user}
\end{figure}

Analyzing the proposals of our team, UNSL\#1 obtained better performance than UNSL\#0 in both tasks. As an illustrative example, Figure \ref{fig:eval-user} shows user evaluation for Task 2, where UNSL\#1 correctly resolved the misclassified users by UNSL\#0. Furthermore, it observes that UNSL\#1 tends to minimize the probabilities of the negative user (Figure \ref{fig:neg-user}) and maximize those of the positive user (Figure \ref{fig:pos-user}). It also shows that UNSL\#1 detected the positive user in post 23 (decision delay=23), a reasonable instance considering the number of user posts. 

Considering the latency-based metrics, our proposals demonstrated satisfactory results, particularly in the ERDE-30 and latency-weighted F1 metrics. However, the results for ERDE-5 were less favorable. It would be interesting to explore potential strategies to enhance the performance of the models for ERDE-5 without compromising the other metrics. This could involve optimizing the classification performance of the models and aligning them with the decision policy proposed in this work. Additionally, it would be worth analyzing new decision policies prioritizing speed and efficiency.

Finally, considering the mean values among all the teams, it is observed that Task 2 was more challenging than Task 1. This fact was probably due to the subjectivity level with which users expressed themselves in each domain, which may have impacted the performance of the models. For example, the post  \emph{``Me gustaría poder comer sin sentir culpa como antes''} (I wish I could eat without feeling guilty like before) could be linked to a user at risk for an eating disorder; however, the text \emph{``Esta semana fue difícil para mí''} (This week was hard for me), it would be rushed to associate it directly with a user with depression.

\section{Conclusion}
\label{sec:conclusion}
In this first edition of the MentalRiskES challenge, our research group solved Tasks 1 and 2. We applied the BETO model by extending its vocabulary with important words, and we used a decision policy based on a historic rule to detect users with depression and eating disorders as early as possible. The method obtained excellent results, demonstrating its effectiveness and consistency in solving these problems in a challenging and underexplored language such as Spanish. 

As future work, the classification models could be refined, analyzing the important words considered to extend the vocabulary, improving the representation of the analyzed instances during user evaluation, and testing the performance of other classification models. Furthermore, it would be interesting to evaluate other decision policies to improve the performance of the models in terms of latency.




\begin{thebibliography}{18}
\expandafter\ifx\csname natexlab\endcsname\relax\def\natexlab#1{#1}\fi
\providecommand{\url}[1]{\texttt{#1}}
\providecommand{\href}[2]{#2}
\providecommand{\path}[1]{#1}
\providecommand{\DOIprefix}{doi:}
\providecommand{\ArXivprefix}{arXiv:}
\providecommand{\URLprefix}{URL: }
\providecommand{\Pubmedprefix}{pmid:}
\providecommand{\doi}[1]{\href{http://dx.doi.org/#1}{\path{#1}}}
\providecommand{\Pubmed}[1]{\href{pmid:#1}{\path{#1}}}
\providecommand{\bibinfo}[2]{#2}
\ifx\xfnm\relax \def\xfnm[#1]{\unskip,\space#1}\fi
\bibitem[{Charlson et~al.(2019)Charlson, van Ommeren, Flaxman, Cornett, Whiteford, and Saxena}]{charlson2019new}
\bibinfo{author}{F.~Charlson}, \bibinfo{author}{M.~van Ommeren}, \bibinfo{author}{A.~Flaxman}, \bibinfo{author}{J.~Cornett}, \bibinfo{author}{H.~Whiteford}, \bibinfo{author}{S.~Saxena},
\newblock \bibinfo{title}{New who prevalence estimates of mental disorders in conflict settings: a systematic review and meta-analysis},
\newblock \bibinfo{journal}{The Lancet} \bibinfo{volume}{394} (\bibinfo{year}{2019}) \bibinfo{pages}{240--248}.
\bibitem[{Aliverdi et~al.(2022)Aliverdi, Farajidana, Tourzani, Salehi, Qorbani, Mohamadi, and Mahmoodi}]{aliverdi2022social}
\bibinfo{author}{F.~Aliverdi}, \bibinfo{author}{H.~Farajidana}, \bibinfo{author}{Z.~M. Tourzani}, \bibinfo{author}{L.~Salehi}, \bibinfo{author}{M.~Qorbani}, \bibinfo{author}{F.~Mohamadi}, \bibinfo{author}{Z.~Mahmoodi},
\newblock \bibinfo{title}{Social networks and internet emotional relationships on mental health and quality of life in students: structural equation modelling},
\newblock \bibinfo{journal}{BMC psychiatry} \bibinfo{volume}{22} (\bibinfo{year}{2022}) \bibinfo{pages}{1--10}.
\bibitem[{Maulik et~al.(2010)Maulik, Eaton, and Bradshaw}]{maulik2010effect}
\bibinfo{author}{P.~K. Maulik}, \bibinfo{author}{W.~W. Eaton}, \bibinfo{author}{C.~P. Bradshaw},
\newblock \bibinfo{title}{The effect of social networks and social support on common mental disorders following specific life events},
\newblock \bibinfo{journal}{Acta Psychiatrica Scandinavica} \bibinfo{volume}{122} (\bibinfo{year}{2010}) \bibinfo{pages}{118--128}.
\bibitem[{Mart{\'\i}nez-L{\'\i}bano et~al.(2022)Mart{\'\i}nez-L{\'\i}bano, Gonz{\'a}lez~Campusano, Pereira~Castillo et~al.}]{martinez2022redes}
\bibinfo{author}{J.~Mart{\'\i}nez-L{\'\i}bano}, \bibinfo{author}{N.~Gonz{\'a}lez~Campusano}, \bibinfo{author}{J.~I. Pereira~Castillo}, et~al.,
\newblock \bibinfo{title}{Las redes sociales y su influencia en la salud mental de los estudiantes universitarios: Una revisi{\'o}n sistem{\'a}tica}  (\bibinfo{year}{2022}).
\bibitem[{Losada and Crestani(2016)}]{losada2016erisk}
\bibinfo{author}{D.~E. Losada}, \bibinfo{author}{F.~Crestani},
\newblock \bibinfo{title}{A test collection for research on depression and language use},
\newblock in: \bibinfo{booktitle}{Proc. of Conference and Labs of the Evaluation Forum (CLEF 2016)}, \bibinfo{address}{Evora, Portugal}, \bibinfo{year}{2016}, pp. \bibinfo{pages}{28--39}.
\bibitem[{Losada et~al.(2017)Losada, Crestani, and Parapar}]{losada2017erisk}
\bibinfo{author}{D.~E. Losada}, \bibinfo{author}{F.~Crestani}, \bibinfo{author}{J.~Parapar},
\newblock \bibinfo{title}{erisk 2017: Clef lab on early risk prediction on the internet: experimental foundations},
\newblock in: \bibinfo{booktitle}{International Conference of the Cross-Language Evaluation Forum for European Languages}, \bibinfo{organization}{Springer}, \bibinfo{year}{2017}, pp. \bibinfo{pages}{346--360}.
\bibitem[{Losada et~al.(2018)Losada, Crestani, and Parapar}]{losada2018overview}
\bibinfo{author}{D.~E. Losada}, \bibinfo{author}{F.~Crestani}, \bibinfo{author}{J.~Parapar},
\newblock \bibinfo{title}{Overview of erisk: early risk prediction on the internet},
\newblock in: \bibinfo{booktitle}{International Conference of the Cross-Language Evaluation Forum for European Languages}, \bibinfo{organization}{Springer}, \bibinfo{year}{2018}, pp. \bibinfo{pages}{343--361}.
\bibitem[{Losada et~al.(2019)Losada, Crestani, and Parapar}]{losada2019overview}
\bibinfo{author}{D.~E. Losada}, \bibinfo{author}{F.~Crestani}, \bibinfo{author}{J.~Parapar},
\newblock \bibinfo{title}{Overview of erisk 2019 early risk prediction on the internet},
\newblock in: \bibinfo{booktitle}{Experimental IR Meets Multilinguality, Multimodality, and Interaction: 10th International Conference of the CLEF Association, CLEF 2019, Lugano, Switzerland, September 9--12, 2019, Proceedings 10}, \bibinfo{organization}{Springer}, \bibinfo{year}{2019}, pp. \bibinfo{pages}{340--357}.
\bibitem[{Losada et~al.(2020)Losada, Crestani, and Parapar}]{losada2020erisk}
\bibinfo{author}{D.~E. Losada}, \bibinfo{author}{F.~Crestani}, \bibinfo{author}{J.~Parapar},
\newblock \bibinfo{title}{erisk 2020: Self-harm and depression challenges},
\newblock in: \bibinfo{booktitle}{Advances in Information Retrieval: 42nd European Conference on IR Research, ECIR 2020, Lisbon, Portugal, April 14--17, 2020, Proceedings, Part II 42}, \bibinfo{organization}{Springer}, \bibinfo{year}{2020}, pp. \bibinfo{pages}{557--563}.
\bibitem[{Parapar et~al.(2021)Parapar, Mart{\'\i}n-Rodilla, Losada, and Crestani}]{parapar2021overview}
\bibinfo{author}{J.~Parapar}, \bibinfo{author}{P.~Mart{\'\i}n-Rodilla}, \bibinfo{author}{D.~E. Losada}, \bibinfo{author}{F.~Crestani},
\newblock \bibinfo{title}{Overview of erisk 2021: Early risk prediction on the internet},
\newblock in: \bibinfo{booktitle}{International Conference of the Cross-Language Evaluation Forum for European Languages}, \bibinfo{organization}{Springer}, \bibinfo{year}{2021}, pp. \bibinfo{pages}{324--344}.
\bibitem[{Parapar et~al.(2022)Parapar, Mart{\'\i}n-Rodilla, Losada, and Crestani}]{parapar2022overview}
\bibinfo{author}{J.~Parapar}, \bibinfo{author}{P.~Mart{\'\i}n-Rodilla}, \bibinfo{author}{D.~E. Losada}, \bibinfo{author}{F.~Crestani},
\newblock \bibinfo{title}{Overview of erisk 2022: Early risk prediction on the internet},
\newblock in: \bibinfo{booktitle}{Experimental IR Meets Multilinguality, Multimodality, and Interaction: 13th International Conference of the CLEF Association, CLEF 2022, Bologna, Italy, September 5--8, 2022, Proceedings}, \bibinfo{organization}{Springer}, \bibinfo{year}{2022}, pp. \bibinfo{pages}{233--256}.
\bibitem[{{Mármol-Romero} et~al.(2023){Mármol-Romero}, {Moreno-Muñoz}, {Plaza-del-Arco}, {Molina-González}, {Martín-Valdivia}, {Ureña-López}, and {Montejo-Ráez}}]{MentalRiskES2023}
\bibinfo{author}{A.~M. {Mármol-Romero}}, \bibinfo{author}{A.~{Moreno-Muñoz}}, \bibinfo{author}{F.~M. {Plaza-del-Arco}}, \bibinfo{author}{M.~D. {Molina-González}}, \bibinfo{author}{M.~T. {Martín-Valdivia}}, \bibinfo{author}{L.~A. {Ureña-López}}, \bibinfo{author}{A.~{Montejo-Ráez}},
\newblock \bibinfo{title}{Overview of {M}entalrisk{ES} at {I}ber{LEF} 2023: {E}arly {D}etection of {M}ental {D}isorders {R}isk in {S}panish},
\newblock \bibinfo{journal}{Procesamiento del Lenguaje Natural} \bibinfo{volume}{71} (\bibinfo{year}{2023}).
\bibitem[{Loyola et~al.(2021)Loyola, Burdisso, Thompson, Cagnina, and Errecalde}]{loyola2021unsl}
\bibinfo{author}{J.~M. Loyola}, \bibinfo{author}{S.~Burdisso}, \bibinfo{author}{H.~Thompson}, \bibinfo{author}{L.~C. Cagnina}, \bibinfo{author}{M.~Errecalde},
\newblock \bibinfo{title}{Unsl at erisk 2021: A comparison of three early alert policies for early risk detection.},
\newblock in: \bibinfo{booktitle}{CLEF (Working Notes)}, \bibinfo{year}{2021}, pp. \bibinfo{pages}{992--1021}.
\bibitem[{Loyola et~al.(2022)Loyola, Thompson, Burdisso, and Errecalde}]{loyola2022unsl}
\bibinfo{author}{J.~M. Loyola}, \bibinfo{author}{H.~Thompson}, \bibinfo{author}{S.~Burdisso}, \bibinfo{author}{M.~Errecalde},
\newblock \bibinfo{title}{Unsl at erisk 2022: Decision policies with history for early classification}  (\bibinfo{year}{2022}).
\bibitem[{Loyola et~al.(2018)Loyola, Errecalde, Escalante, and Montes~y Gomez}]{loyola2018learning}
\bibinfo{author}{J.~M. Loyola}, \bibinfo{author}{M.~L. Errecalde}, \bibinfo{author}{H.~J. Escalante}, \bibinfo{author}{M.~Montes~y Gomez},
\newblock \bibinfo{title}{Learning when to classify for early text classification},
\newblock in: \bibinfo{booktitle}{Computer Science--CACIC 2017: 23rd Argentine Congress, La Plata, Argentina, October 9-13, 2017, Revised Selected Papers 23}, \bibinfo{organization}{Springer}, \bibinfo{year}{2018}, pp. \bibinfo{pages}{24--34}.
\bibitem[{Devlin et~al.(2018)Devlin, Chang, Lee, and Toutanova}]{devlin2018bert}
\bibinfo{author}{J.~Devlin}, \bibinfo{author}{M.-W. Chang}, \bibinfo{author}{K.~Lee}, \bibinfo{author}{K.~Toutanova},
\newblock \bibinfo{title}{Bert: Pre-training of deep bidirectional transformers for language understanding},
\newblock \bibinfo{journal}{arXiv preprint arXiv:1810.04805}  (\bibinfo{year}{2018}).
\bibitem[{Cañete et~al.(2020)Cañete, Chaperon, Fuentes, Ho, Kang, and Pérez}]{CaneteCFP2020}
\bibinfo{author}{J.~Cañete}, \bibinfo{author}{G.~Chaperon}, \bibinfo{author}{R.~Fuentes}, \bibinfo{author}{J.-H. Ho}, \bibinfo{author}{H.~Kang}, \bibinfo{author}{J.~Pérez},
\newblock \bibinfo{title}{Spanish pre-trained bert model and evaluation data},
\newblock in: \bibinfo{booktitle}{PML4DC at ICLR 2020}, \bibinfo{year}{2020}.
\bibitem[{Burdisso et~al.(2019)Burdisso, Errecalde, and Montes-y G{\'o}mez}]{burdisso2019text}
\bibinfo{author}{S.~G. Burdisso}, \bibinfo{author}{M.~Errecalde}, \bibinfo{author}{M.~Montes-y G{\'o}mez},
\newblock \bibinfo{title}{A text classification framework for simple and effective early depression detection over social media streams},
\newblock \bibinfo{journal}{Expert Systems with Applications} \bibinfo{volume}{133} (\bibinfo{year}{2019}) \bibinfo{pages}{182--197}.

\end{thebibliography}
\end{document}